Sejong Face Database: Multi-Modal Disguised Face Database
Usman Cheema and Seungbin Moon*

*Abstract*: Commercial application of facial recognition demands robustness to a variety of challenges such as illumination, occlusion, spoofing, disguise etc. Disguised face recognition is one of the emerging issues for access control systems, such as security check points at the borders. However, lack of availability of face databases, with a variety of disguise addons, limits the development of academic research in the area. In this paper, we present a multimodal disguised face dataset to facilitate the disguised face recognition research. The presented database contains 8 facial addons and 7 additional combinations of these addons to create a variety of disguised face images. Each facial image is captured in visible, visible plus infrared, infrared, and thermal spectra. Specifically, the database contains 100 subjects divided into subset-A (30 subjects, 1 image per modality) and subset-B (70 subjects, 5 plus images per modality). We also present baseline face detection results performed on the proposed database to provide reference results and compare the performance in different modalities. Additionally, qualitative and quantitative analysis is performed to evaluate the challenging nature of disguise addons. The dataset will be publicly available with the acceptance of the research article.

1. Introduction

Facial recognition in a commercial system should address a wide range of imperfections such as illumination, pose, expression, aging, obstruction, noncooperation, and facial disguise. With the emerging implementation of automated facial recognition systems at security sensitive areas, disguise is becoming one of the greatest challenges for facial recognition [1]. Disguise addons, such as wigs, masks, and makeup can easily alter one's appearance to deceive facial recognition systems. Various algorithmic approaches have been proposed against deception and disguises, such as holistic [2], [3], local [4], and patch based [5] methods. We refer the reader to survey papers [6], [7] for detailed reading.

Infrared, thermal, and spectrum imaging has shown promise against disguise and other challenges of facial recognition [3]. Infrared modality has shown advantages over visible modality, especially against illumination and disguise addons [8]. Although, an infrared light source is required, infrared performs well against illumination as it can capture images at extremes of visible light conditions. Human skin and hair exhibit am infrared signature unique from disguise addons, e.g., mask, eyeglasses. These addons can be differentiated from skin and hair using appropriate sensors and knowledge of these signatures.

Thermal imagery is an emerging modality, that is gaining more attention due to the introduction of affordable microbolometer sensors. As thermal imaging relies on the heat radiation emitted by the objects themselves, it does not require natural or artificial light sources and therefore is invariant to illumination changes. Disguise addons, as they are not part of human body, exhibit a lower temperature than human skin and thus appear darker than human skin or hair. Additionally, subcutaneous anatomical information, such as vascular network and tissue density can be extracted from thermal images as a separate modality [9].

Multi modal disguised face datasets play a key role in training and evaluation of these algorithms. Large scale databases have been recently presented for disguised faces [10] recognition, but as they are collected through publicly available celebrity images, they lack disguise labels and are limited to visible modality. Primarily, databases collected in visible infrared, visible depth [11], and visible thermal modalities [12], lack variations in disguise addons as they were collected with a focus on non-disguise aspects of facial recognition. Thus, these databases offer insufficient variations for disguise detection. We summarize the currently available multi modal disguise face databases and analyze their advantages and disadvantages for disguise recognition. I²BVSD [5], [13] contains images in visible and thermal spectra with variations in fake facial hair, caps, wigs, masks, glasses, and expressions. The database contains images of 75 subjects, 15 females and 60 males, of south Asian ethnicity. The images are captured under constant illumination with neutral expression and frontal pose. The disguise variations included in the database are variations in hair styles, beard and mustache, glasses, cap, mask and combinations of these disguise accessories, however no labeling information is provided for the disguise addons. Each subject has at least one neutral face image and at 5 to 9 frontal disguised images, amounting to a total of 681 images for each modality.

The BRSU Spoof Database [14], [15] contains images captured in the visible and infrared modalities at four different spectral frequencies (935 nm, 1060 nm, 1300 nm, and 1550 nm). The database contains variations in expression, makeup, 3D masks, fake beard, glasses, fake nose and presentation attack. The database contains images of 5 subjects with 9 to 30 disguise addons for each subject. The images are captured using a SWIR camera sensitive to a spectral range of 900-1700 nm with frequency specific LED ring light source for face illumination. The capturing setup used to collect BRSU spoof database is sensitive to relative position of the subject, illumination source, subject movement and external illumination. Thus, fixed pattern noise correction, motion compensation and calibration has been employed for image preprocessing.

Spectral Disguise Face Dataset [16] contains images in visible and eight narrow bands across the near infrared (530nm to 1000nm) spectrum. The subjects contributing to this database are all male participants between the age of 20 to 50 years. The data collected is categorized into bona fide set (normal captured samples without disguise), and two variants of disguise that includes wearing normal beard and long beard. The subjects in the bona fide presentation have a naturally grown moustache and 22 out of 54 subjects have a naturally grown beard. In each session, 5 sample images per subject are collected for 3 sessions, which corresponds to 54 subjects × 5 samples × 8 bands × 3 sessions = 6480 samples.

TABLE I
COMPARISON OF AVAILABLE DISGUISE FACE DATABASES

| Data Base | No. of Subjects | No. of Images | Gender M:F | Visible | Infrared | Thermal | Visible plus Infrared | No. of Addons | Combination Addons |
|---|---|---|---|---|---|---|---|---|---|
| I²BVSD | 75 | 1,362 | 60:15 | ✓ | | ✓ | | 5 | ✓ |
| BRSU | 5 | 660 | 4:1 | ✓ | ✓ | | | 4-12 | ✓ |
| SDFD | 54 | 6,480 | 54:0 | ✓ | ✓ | | | 3 | |
| Proposed-A | 30 | 1,800 | 16:14 | ✓ | ✓ | ✓ | ✓ | 13 | ✓ |
| Proposed-B | 70 | 23,000 | 44:26 | ✓ | ✓ | ✓ | ✓ | 13 | ✓ |

A practical FR system needs to be cost effective, robust against various challenges of FR, scalable and practical for realistic scenarios. The goal of the study is to provide a practical facial disguise database to facilitate development towards more robust FR systems. With this in mind, we employ a generic imaging system which is easy to install, easy to operate and can provide quality images for a wide range of commercial applications. The details of our image capturing setup are giving in Section II. A wide array of conceivable real-world disguise addons and their combinations are included in the proposed database, amounting to a total of 13 facial addon variations. Moreover, to facilitate directed research, images are labeled such that the database contains precise information for image contents, which is hard to achieve for databases collected from web. To carter to different options available for both genders, gender specific addons are also included. A diverse ethnic variety and gender balance in subjects is achieved to account for a wide variety of skin tones.

The proposed database contains images captured in four modalities, visible, infrared, thermal, and visible plus infrared (VisIr). These modalities, combined, can capture a wide array of features. Comparison of the available disguise databases and the proposed database is presented in Table I. Since our work focuses on multispectral facial addons as form of disguise, hence presentation and replay attack databases and research works are omitted from our summary overview.

In Section II, we provide detailed description of the presented database, technical specification of the capturing apparatus, and design layout of the capturing environment. In Section III, baseline face detection and facial recognition is performed on the proposed database to provide benchmarking results. Furthermore, we present a multimodal fusion setup to improve accuracy over single modal face recognition. The conclusion and future work are presented in the final section.

2. Database Description

2.1 General Characteristics

The proposed database is divided into two subsets, subset-A and subset-B, subset-A being collected one year prior to subset-B. Subset-A contains facial images of 30 subjects, 16 males and 14 females. One neutral and one addon, for each addon, image was captured in each modality. All images are captured with frontal faces. Subset-B contains facial images of 70 subjects, 44 males and 26 females. 15 neutral face images and 5 addon images for remaining addons were captured in each modality. Additional 5 images with real beard for males and makeup for females were also captured. For images collected in the subset-B, minor pose variations were also included. The first image is captured with the subject looking directly at the camera. The second, third, fourth, and fifth images are captured with the subject looking right, left, up, and down, respectively. The degree of movement was controlled by requesting the subjects to look straight at marked points in space, calibrated to induce a 15º rotation. Each subsequent set of 5 images for neutral, real beard, and makeup images were captured repeating the same head movement protocol.

The images for each subject were captured in two sessions, two weeks apart. For the males, neutral face images with shaved facial hair and addons were captured in the first session. In the second session, images were taken after the subjects grew facial hair over a two week period. For the females, neutral face and addon images were captured in the first session. In the second session, images with makeup were captured. Male and female faces were captured with 13 and 12 disguise addons, respectively. As means of disguise may vary between genders, we accommodate this by using different disguise variants. Male subjects were captured with false mustaches and false beards, while female subjects were captured with wig and makeup. Common facial accessories, such as masks, glasses, and caps, were used for both genders. The list of addons and their combinations provided in the database is given in Table II. The accuracy of facial detection and recognition algorithms may vary across ethnicities due to differences in facial features and skin tone. Subjects from various ethnicities were included in the database to include variations in skin tone and facial structure as shown in Fig. 1.

For an easier understanding and usability of the database, each image is named so that its contents can be recognized by its label. Each image is given a name of the format *nu_xy_z.jpg*, where *nu* represents a two digit number unique to each subject, *xy* represents the label for the addon contained in the image as shown in Table II, and *z* represents the image modality of the image, where *v* is for visible, *i* is for visible plus infrared, *d* is for infrared, and *t* is for thermal modality.

TABLE II
LIST OF INCLUDED ADDONS, NUMBER OF IMAGES AVAILABLE PER ADDON, AND LABEL NAME FOR THE IMAGE FILES.

| Add-on | | Label | No. of Images | Gender | |
|---|---|---|---|---|---|
| | | | | Male | Female |
| No Addon | Natural Face | NF | 15 | ✓ | ✓ |
| | Read Beard | RB | 10 | ✓ | |
| Accessory Addon | Cap | CP | 5 | ✓ | ✓ |
| | Scarf | SC | 5 | ✓ | ✓ |
| | Glasses | GL | 5 | ✓ | ✓ |
| | Mask | MA | 5 | ✓ | ✓ |
| | Makeup | MU | 5 | | ✓ |
| Fake Addon | Wig | WI | 10 | | ✓ |
| | Fake Beard | FB | 5 | ✓ | |
| | Fake Mustache | FM | 5 | ✓ | |
| Combination Addon | Wig-Glasses | WG | 5 | | ✓ |
| | Wig-Scarf | WS | 5 | | ✓ |
| | Cap-Scarf | CS | 5 | ✓ | ✓ |
| | Glasses-Scarf | GS | 5 | ✓ | ✓ |
| | Glasses-Mask | GM | 5 | ✓ | ✓ |
| | Fake Beard-Cap | BC | 5 | ✓ | |
| | Fake Beard-Glasses | BG | 5 | ✓ | |

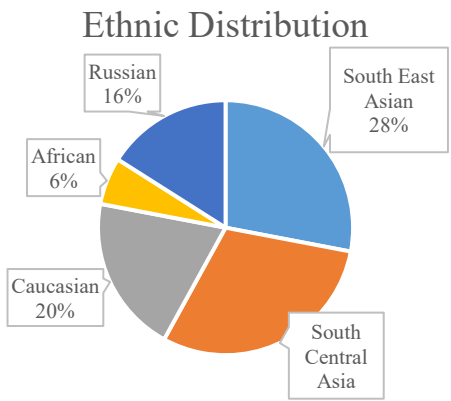

Fig. 1. The ethnic distribution of the subjects in the database.

2.2. Imaging Hardware and Capturing Environment

Three imaging cameras were employed to capture face images in four modalities. A smart phone camera was used to capture images in the visible modality. Images in visible modality were captured in fluorescent illumination, with four area-type light sources fixed on the ceiling illuminated the scene. Raspberry Pi NoIR, a color imaging camera with the infrared filter removed, was used to capture images in infrared and infrared plus visible modality. Pi NoIR can capture images in 400 nm to 1,400 nm wavelength range of the electromagnetic spectrum. The standard infrared module shipped with Pi NoIR was used to illuminate the subjects face from front to capture the infrared images. All visible light sources in the environment were either turned off or blocked while capturing the infrared images. Visible plus infrared images were captured in the presence of visible and infrared illumination source. Therm-App was used to capture thermal images, it uses 750 nm to 1,400 nm wavelength range of electromagnetic spectrum. The specifications of imaging devices are summarized in Table III.

The three imaging devices were mounted close to each other to achieve a similar viewing angle of the subject's face. The camera setup was mounted on a tripod stand and positioned two meters from the subject at eye level as shown in Fig. 2. The smart phone and Therm-App cameras were controlled through a common application, allowing in sync image capturing of visible and thermal images. The images captured through Raspberry Pi NoIR had a slight delay, but the effects of delay were minimized by requesting the subjects to remain stationery.

TABLE III
SPECIFICATIONS OF IMAGING DEVICES

| Modality | Camera | Illumination | Resolution (px) | Frequency |
|---|---|---|---|---|
| Visible | Smart Phone | Fluorescent | 4,032 × 3,024 | 400 nm to 770 nm |
| Infrared | Pi NoIR | Infrared | 1,680 × 1,050 | 770 nm to 1,000 nm |
| Ir-Vis | Pi NoIR | Fluorescent and Infared | 1,680 × 1,050 | 400 nm to 1,000 nm |
| Thermal | Therm-App | ~ | 768 × 756 | 750 nm to 1,400 nm |

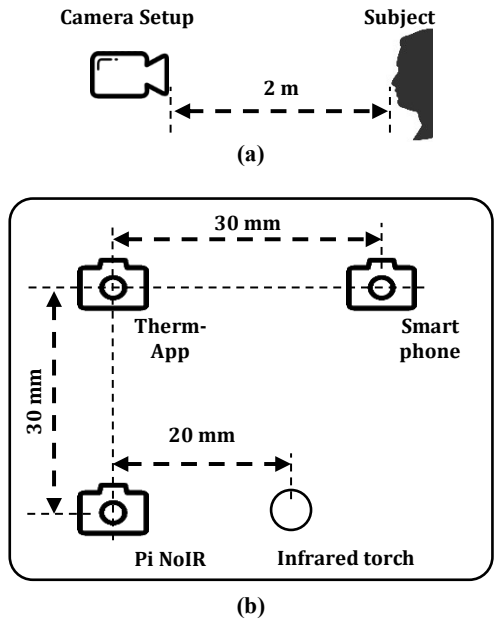

Fig. 2. Setup of imaging equipment. (a) Relative positioning of the camera from the subject. (b) Relative arrangement of the cameras.

The ambient temperature of the capturing facility was maintained at 25 ± 5°C to eliminate environmental interference with thermal imaging. Additionally, subjects were asked to spend at least ten minutes in the same environment prior to the image capturing process. This request was made so that the subject's body could come into equilibrium with the environment and any effects from prior activities and surroundings should be minimized. Sample images from subset-B for one subject in four modalities are shown in Fig. 3.

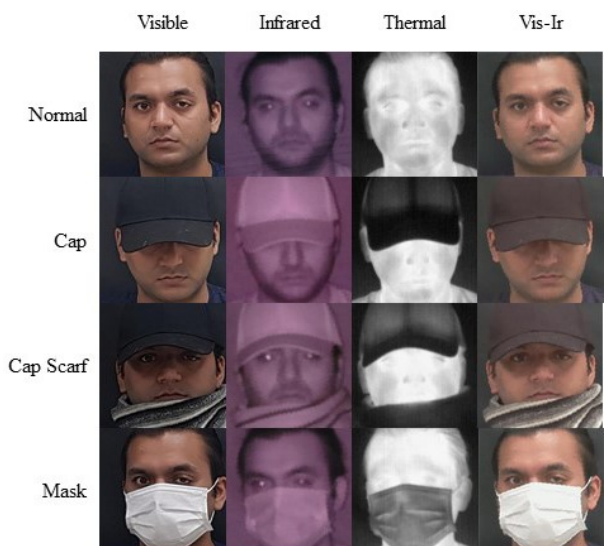

Fig. 3. Sample images of from one subject.

3. Benchmark Evaluations

In this section, we evaluate the quality of the proposed database by performing baseline face detection and face recognition experiments. The protocols used have been tested on publicly available face databases, and we compare results on the presented database.

3.1 Image preprocessing for face detection

Each camera having its own spatial resolution, the images were captured in different resolutions as shown in Table III. Inherently, this results in different face sizes in each image. As first step of preprocessing we calculated the magnification factor for smaller images such that the face size in infrared, VisIr, and thermal images matches the size of face in visible images. The magnification was calculated based on four reference points in the background captured by all 3 cameras. The smaller images were magnified such that all images contain the same face size. Next, the quartet of images, one for each modality, were aligned using the reference points from the background. The validity of magnification and alignment process was manually confirmed on randomly selected subsample images from each subject. A 1380 × 1061 crop was performed on the resized and aligned images to remove background information. The resulting images contain face size in the range of 300px to 600px.

3.2 Face Detection

Face detection and image processing are key components of a robust facial recognition system. Facial disguises such as cap and scarf hide key landmarks of the face, lowering face detection accuracy. Benchmarking face detection was performed using OpenCV's implementation of Viola Jones face detector [17], OpenCV's neural network face detector based on "Single-Shot-Multibox detector" (SSMD) [18] which uses ResNet-10 [19] architecture as backbone, Dlib's Histogram of Oriented Gradients [20] fed into an SVM based face detector [21], and Dlib's implementation of Maximum-Margin Object Detector (MMOD) [22] for face detection. Face detection was performed on subset-B of the proposed dataset. High detection results have been reported [23–25] using these algorithms on faces without disguise addons. Face detection accuracy was measured for each addon (total 17 addons), in each modality (4 modalities). For simplification and ease of understanding we classify the addons into three categories, easy, medium, and hard, based on the average performance of our face detectors for each class as shown in Table IV.

Face detection accuracies of visible, infrared, visible plus infrared, and thermal images are presented in Fig. 4. Furthermore, we present the accuracy of each algorithm averaged over visible, infrared, and VisIr modalities. Results from thermal modality are excluded from Fig. 4. (b) due to large modality gap between training and testing data of the employed algorithms.

**Accuracy across disguise.** Face detection relies on key facial features and lank marks to identify faces in the image. The presence of disguise addons can obstruct facial features, lowering face detection accuracy. We observe that an average of below 60% face detection accuracy is achieved for cap, cap scarf, fake beard cap, glasses mask and, scarf glasses, these are classified as *hard* addons. Cap is worn by the subjects such that the eyes are hidden, obstructing the eyes having the largest impact on face detection. Mask is worn such that the lower face region including the nose is covered, that combined with glasses provides and effective disguise for face detection and subsequently facial recognition. 82% to 91% face detection accuracy is achieved for fake beard, fake beard glasses, glasses, mask, scarf, wig glasses, and wig scarf, these are classified as *medium* addons. As these addons cover relatively lesser portion of the face, we observe an improved accuracy over *hard* addons. 98% to100% accuracy is observed for normal, real beard, fake mustache, makeup, and wig, making them the least of the troublesome forms of disguise as far as face detection is concerned. MMOD and SSMD achieve 100% face recognition accuracy against real beard, fake mustache, makeup, and wig in visible, infrared and infrared plus visible modalities.

TABLE IV
DISGUISE ADDON CATEGORIZATION BASED ON AVERAGE PERFORMANCE OF DETECTORS ACROSS ALL MODALITIES

| Category | Easy | Medium | Hard |
|---|---|---|---|
| Average Accuracy | Top 30% | 30% to 70% | Bottom 30% |
| Addon Type | Normal Beard Fake Mustache Makeup Wig | Fake Beard Fake Beard-Glasses Glasses Mask Scarf Wig-Glasses Wig-Scarf | Cap Cap-Scarf Fake Beard-Cap Glasses-Mask Scarf-Glasses |

**Accuracy in different modalities.** Pretrained face detection modules used in our work rely on visible images for training. While, infrared and visible plus infrared modalities mimic visible modality closely, thermal imagery captures heat signature of the face, which vary significantly from visible modality. This difference is reflected in our experimental results as the average accuracy of face detection suffers greatly in thermal modality. SSMD achieves highest average accuracy of 95% in visible plus infrared modality followed by visible modality at 94%, proving little effectiveness of employing visible plus infrared over visible modality in a well illuminated environment. 88% facial detection accuracy is achieved in infrared modality, proving its effectiveness in dark environments. An overall highest accuracy of 95% is achieved using SSMD detector in visible plus infrared modality. While reliable accuracy is achieved in the visible and infrared modalities, modality specific algorithm needs to be employed for face detection in thermal modality.

3.3. Disguised Face Recognition

Face recognition is performed on the presented datasets using Squeeze and Excitation blocks [26] using ResNet backbone as presented in VGGFace2 [27]. The network has proven to be effective at recognizing faces across pose, age, illumination, and makeup. The network is trained for 3 settings: (a) pre trained network weights are used from VGGFace2 dataset [27] and networks are trained further on each modality independently (SeNet-Trained, Se-Tr for short), no layers are frozen, (b) pre trained network weights are used from VGGFace2 and all but the last pooling layer are frozen (SeNet-Frozen, Se-Fr for short), (c) same as (b) and score fusion is performed for the four modalities (SeNet-Fused, Se-Fu for short).

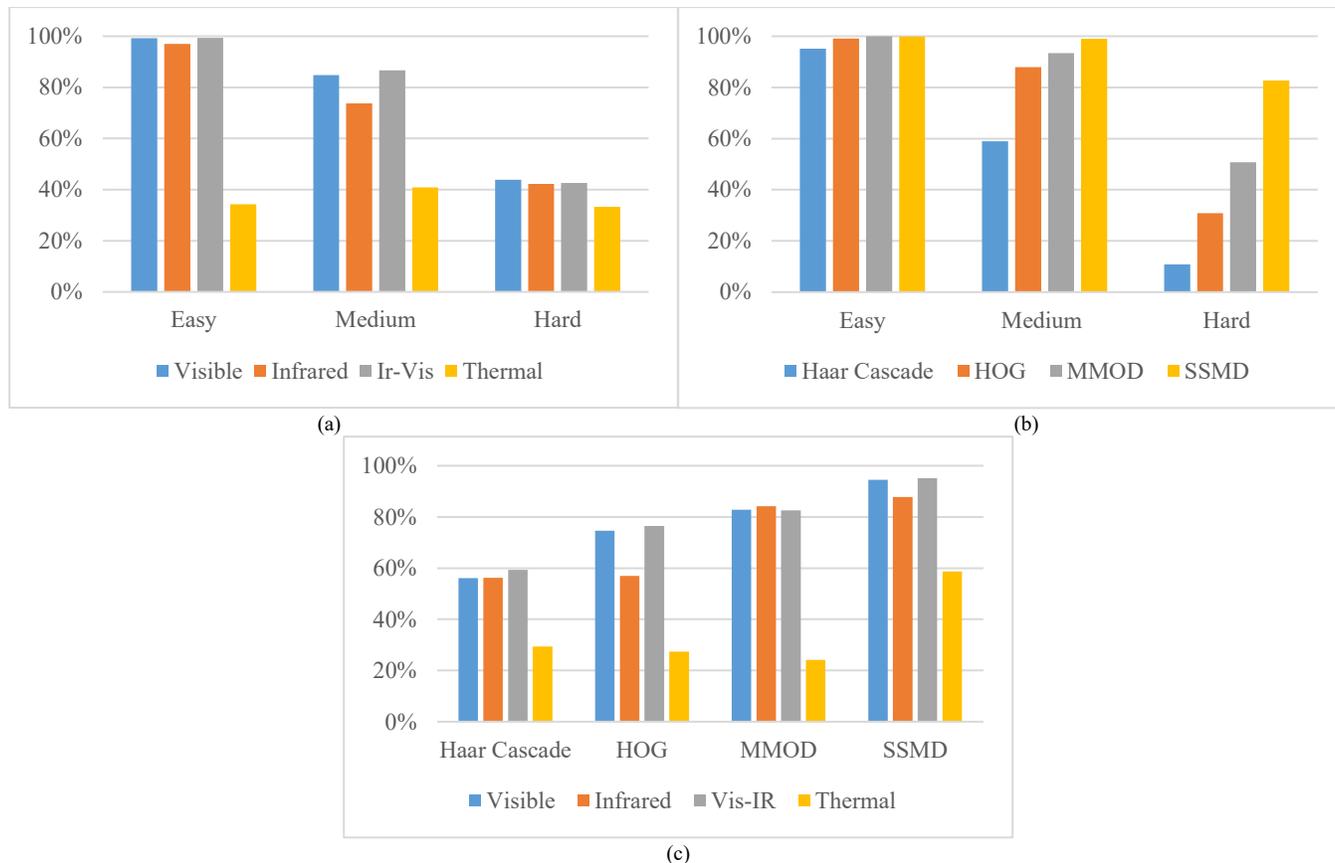

Fig. 4. Face detection results on the proposed dataset. (a) Average face detection accuracy of Haar Cascade, HOG-SVM, MMOD, and SSMD for easy, medium, and hard categories of facial disguises. (b) Individual accuracy of Haar Cascade, HOG SVM, MMOD, and SSMD averaged over visible, infrared, and visible plus infrared modalities. (c) Accuracy of four face detection algorithms averaged over all facial addons in each modality.

SSMD face detector is used to remove redundant background data from face images. SSMD achieves an overall accuracy of 95% or VisIr images (Section 3.2). As we already have scaled, aligned, and cropped images as described earlier, we expect the face data to be in the similar spatial location across a quadruple of images in four modalities. Face detection is performed on each image in the quadruple, if the detection is valid, the face region of interest (ROI) is retained. The retained ROI from each image are merged to generate a super face region. Super region is then transformed into a square shape containing the cropped face. Quadruples where not a single face was detected in any modality are not used for face recognition.

The experiments are performed on two different training data distributions: (a) Normal Training: 9 normal face images per subject; (b) Base Addon Training: 4 normal and 14 disguised face images per class. Face images captured with one facial disguise addon are included in the base addon training data. The included addons are cap, scarf, fake beard, fake mustache, glasses, mask and scarf. This is to test the performance of the networks trained on disguised face images compared to a network trained on normal face images.

TABLE V
DETAILS OF TRAINING AND TEST DATA SPLITS FOR FACE RECOGNITION

|  | Normal | Disguise | Augmented | Subjects | Total |
|---|---|---|---|---|---|
| Train: Normal | 9 | 0 | 16 | 70 | 10,080 |
| Train: Base Addon | 4 | 14 | 16 | 70 | 20,160 |
| Test | 3 | 38 | 0 | 70 | 7,980 |

Additionally, data augmentation is performed on the training data to increase variation and avoid overfitting during the training process. Training images are vertically flipped, rotated (-15º and 15º), shifted (x-axis and y-axis for -20 px and 20 px), and scaled (factor of 0.8 and 1.2) to generate a larger training size. The detailed breakdown of train and test data is given in Table V.

**Facial recognition.** Face recognition aims to predict, for a given test image, whose face it is. For each of the 70 subjects in subset-B, 41 images are used for testing as shown in Table V. The networks are trained independently for each modality and tested on the same modality. A top-1 classification error, for the imaging modality with lowest error, is used to evaluate the accuracy of the tested networks on the proposed dataset as shown in Table. VI. To evaluate the complexity of the proposed dataset we compare our results to SeNet architecture trained and tested on VGGFace2 dataset. SeNet architecture achieves a 3.9% error rate on VGGFace2 dataset, the data set contains 3.31M images of 9,131 subjects with variations in pose, age, illumination and makeup. Table VI shows the top-1 error rates as reported in [27] and our experiments.

**Training on normal faces vs addon faces.** Networks *(a)* and *(c)* are trained on 9 normal face images per subject. Networks *(b)* and *(d)* are trained on 3 normal and 38 disguised face images, only single addons are used for disguised face images. Significant improvement in classification is achieved when the networks are trained on normal plus base addon images (networks *(a)* and *(c)* compared to networks *(b)* and *(d)*. The addition of base addons for training allow the networks to learn variations in facial information, and partial occlusion without over training the network.

For all the presented setups, the pretrained network weights on VGGFace2 were employed. Networks *(a)* and *(b)* were fine tuned with the proposed dataset with no frozen layers, while networks *(c)* and *(d)* were fine tuned with all but the last pooling layer frozen. Freezing the convolution layers results in a further decrease in error of 5% for individual modalities, (network *(b)*, compared to network *(d)*.

Table VI
ERROR OF SENET WITH RESNET-50 BACKBONE ON THE PROPOSED DATASET.

| Network Name | Network Architecture | Pretrain Dataset | Training Dataset | Test Dataset | Error (%) |
|---|---|---|---|---|---|
| VGGFace | ResNet-50 | VGGFace | VGGFace2 | VGGFace2 | 10.6 (Visible) |
| MS1M | ResNet-50 | MS-Celeb-1M | VGGFace2 | VGGFace2 | 5.6 (Visible) |
| VGGFace2 | SeNet | Scratch | VGGFace2 | VGGFace2 | 3.9 (Visible) |
| (a) Se-Tr | SeNet | Scratch | Proposed Normal | Proposed | 65 (Visible) |
| (b) Se-Tr | SeNet | Scratch | Proposed Base Addon | Proposed | 28 (Vis-Ir) |
| (c) Se-Fr | SeNet | VGGFace2 | Proposed Normal | Proposed | 39 (Visible) |
| (d) Se-Fr | SeNet | VGGFace2 | Proposed Base Addon | Proposed | 23 (Visible) |
| (e) Se-Fu-Tr | SeNet | VGGFace2 | Proposed Base Addon | Proposed | 17 (Fused) |
| (f) Se-Fu-Fr | SeNet | VGGFace2 | Proposed Base Addon | Proposed | 18 (Fused) |
| (g) e and f fused | SeNet | VGGFace2 | Proposed Base Addon | Proposed | 7.7 (Fused) |

**Multi modal score fusion.** We pretrained eight base networks on VGGFace2 dataset and finetuned them on the proposed dataset with base addon subset. Four of the networks are finetuned with no frozen layers, network *(e)*, the other four are trained with all but the last average pooling layers frozen, network (f). Each of the networks in the two subsets of four networks is trained on different modality. A decreased error of 17% for network *(e)* and 18% for network *(f)* is achieved using multimodal fusion. Score fusion for *(e)* and *(f)*, network *(g)*, achieves an overall face recognition accuracy of 92.3% across all facial addons, an improved accuracy over using a single modality. The detailed results for each addon for network *(e)*, *(f)* and *(g)* are presented in Table VII. The roc plots for each addon for network *(f)* are shown in Fig.4.

Table VII
RECOGNITION ACCURACY OF TESTED NETWORKS ACROSS FACIAL ADDONS IN THE PROPOSED DATABASE.

| Addons | Network (e) | Network (f) | Fused (e), (f) |
|---|---|---|---|
| Fake Beard-Cap | 54.9% | 43.4% | 67.1% |
| Wig-Glasses | 39.6% | 73.0% | 74.2% |
| Cap-Scarf | 68.7% | 55.7% | 77.5% |
| Wig-Scarf | 44.3% | 84.8% | 81.4% |
| Wig | 46.2% | 93.9% | 84.1% |
| Makeup | 74.0% | 94.7% | 94.8% |
| Scarf-Glasses | 95.4% | 68.5% | 95.8% |
| Glasses-Mask | 98.7% | 40.2% | 96.5% |
| Fake BeardGlasses | 98.4% | 54.5% | 98.0% |
| Beard | 87.3% | 99.8% | 99.7% |
| Cap | 99.5% | 92.5% | 100.0% |
| Fake Beard | 100.0% | 99.2% | 100.0% |
| Fake Mustache | 100.0% | 100.0% | 100.0% |
| Glasses | 100.0% | 95.3% | 100.0% |
| Mask | 100.0% | 96.6% | 100.0% |
| Normal | 100.0% | 100.0% | 100.0% |
| Scarf | 100.0% | 99.7% | 100.0% |
| Average | 82.8% | 81.9% | 92.3% |

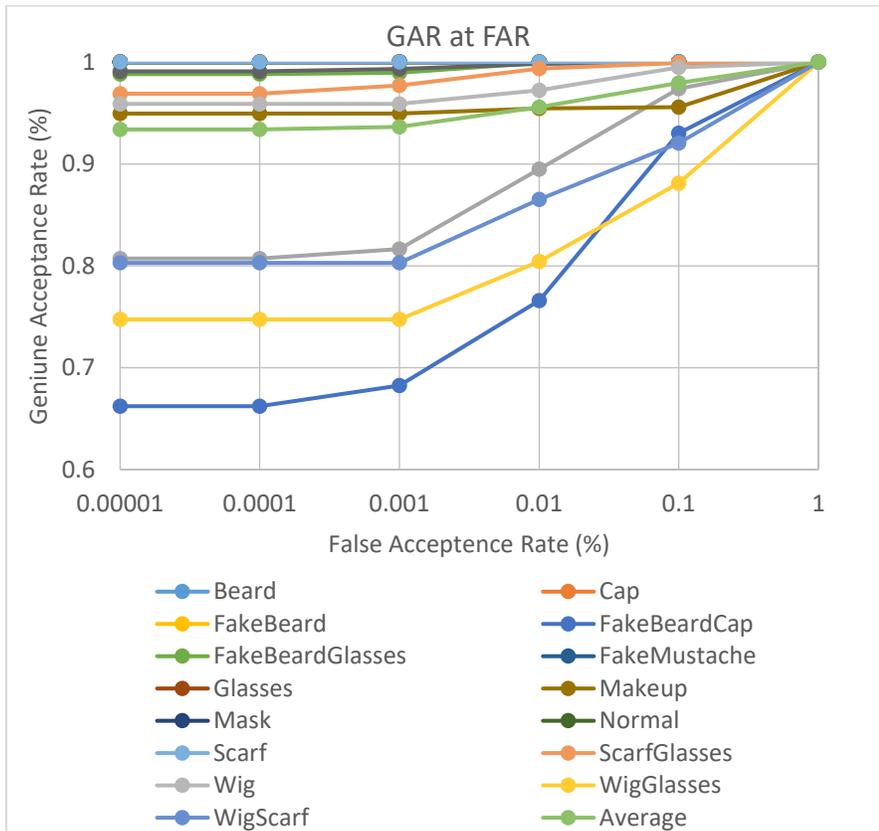

4. Conclusion and Future Work

As disguise addons can easily deceive state of the art facial recognition systems, adequate databases are needed to develop effective disguised facial recognition algorithms. The contribution of this research is to present a practical, cost effective, facial disguise database to facilitate research in the domain of disguised face recognition. The database includes a wide array of facial addons, ethnic variety, and a reasonable gender balance. Moreover, to facilitate directed research, images are labeled such that the database contains precise information for image contents. The database is captured in four imaging modalities, taking advantage of unique features from each modality.

Our baseline experimental results using Squeeze and Excitation networks verify the challenging nature of the presented database. Our proposed setup *(e)* shows the usefulness of the proposed database for deep neural networks. Furthermore, the advantage of using multi modal data is shown. In the future, we will exploit multimodal feature fusion and improved network architectures to further improve the performance.